\documentclass[conference]{IEEEtran}
\IEEEoverridecommandlockouts
\usepackage{cite}
\usepackage{amsmath,amssymb,amsfonts}
\usepackage{algorithmic}
\usepackage{graphicx}
\usepackage{textcomp}
\usepackage{xcolor}
\usepackage{soul}
\usepackage{multirow}
\def\BibTeX{{\rm B\kern-.05em{\sc i\kern-.025em b}\kern-.08em
    T\kern-.1667em\lower.7ex\hbox{E}\kern-.125emX}}
\begin{document}

\title{Progress Notes Classification and Keyword Extraction using Attention based Deep Learning Models with BERT}

\author{
\IEEEauthorblockN{Matthew Tang}
\IEEEauthorblockA{Departments of Statistics and CS \\
University of Illinois at Urbana-Champaign\\
mt13@illinois.edu}
\and
\IEEEauthorblockN{Priyanka Gandhi}
\IEEEauthorblockA{Department of Computer Science \\
IUPUI\\
prgandh@iu.edu}
\and
\IEEEauthorblockN{Md Ahsanul Kabir}
\IEEEauthorblockA{Department of Computer Science \\
IUPUI\\
mdkabir@iu.edu}
\and

\IEEEauthorblockN{Christopher Zou}
\IEEEauthorblockA{Carmel High School\\
zou.christopher@gmail.com}
\and

\IEEEauthorblockN{Jordyn Blakey}
\IEEEauthorblockA{Department of Computer Science \\
DePauw University\\
jordynblakey1@gmail.com}
\and
\IEEEauthorblockN{Xiao Luo}
\IEEEauthorblockA{Department of Computer Information Technology\\
IUPUI\\
luo25@iupui.edu}
}

\maketitle

\begin{abstract}
Various deep
learning algorithms have been developed to analyze different types of clinical data including clinical text classification, and extracting information from ‘free text’ and so on. However, automate the keyword extraction from the clinical notes is still  challenging. The challenges includes dealing with noisy clinical notes which contains various abbreviations, possible typos and unstructured sentences. The objective of this research is to investigate the attention-based deep learning models to classify the de-identified clinical progress notes extracted from a real-world EHR system. The attention-based deep learning models can be used to interpret the models and understand the critical words that drive the correct or incorrect classification of the clinical progress notes. The attention-based models in this research are capable of presenting the human interpretable text classification models. The results show that the fine-tuned BERT with attention layer can achieve a high classification accuracy of 97.6\%, which is higher than the baseline fine-tuned BERT classification model. In this research, we also demonstrate that the attention-based models can identify
relevant keywords that are strongly related to the clinical progress note categories.
\end{abstract}

\section{Introduction}
Deep learning algorithms have been applied to different tasks of text mining and natural language processing, such as identifying parts of speech \cite{li2014recursive} \cite{kabir2016deep}, entity extraction \cite{collobert2008unified} \cite{santos2015boosting}, sentiment analysis \cite{zhang2018deep}, text classification \cite{howard2018universal}, and other aspects of text \cite{chatterjee2019understanding}. In recent years, applications of deep learning and text mining algorithms to the medical data have gained a lot of attention.  Researches have been done on making use of EHR clinical notes for clinical decision support. Typically, the `free-text' clinical notes include discharge summaries, patient instructions and progress notes, which contain patients medical history, family history, treatment history, and so on. Managing and extracting key words or information from the clinical notes by using learning algorithms are always challenging. 

In this research, we develop attention-based deep learning models for classifying a set of clinical progress notes which belongs to 12 different clinical categories. These progress notes are extracted from a large institutional health care center. We build attention-based deep learning models to classify the progress notes. The models are tested on their ability to classify the progress note to the corresponding categories based on their content. Most of the deep learning models require a large amount of training data. We develop our system through making use of the word or token embedding from pre-trained models trained on extensive text collections. To investigate how the attention-based deep learning models perform for progress notes classification, we train and evaluate the attention based approach with several deep learning models, including the most recent language model BERT \cite{devlin2018bert}, and a bidirectional long short-term memory (BiLSTM) model \cite{huang2015bidirectional}. The results show that the BERT model with an additional attention layer can achieve a high classification accuracy of 97.6\%, which is higher than the base fine-tuned BERT classification model.

Typical deep neural networks or machine learning algorithms perform as ``black boxes''. It is hard to understand and interpret the process of decision making. A recent research demonstrated a probe sentence embedding models to interpret the neural network's features on analyzing the clinical notes\cite{ormerod2019analysing}. Different from this approach, we believe the attention-based deep learning models have a built-in mechanism which can be used to identify the keywords that drive the neural network to predict a clinical progress note into a clinical category. We investigate both token embedding and word embedding for attention weight calculation to extract the keywords for interpreting classification reasoning. To visualize the keywords for reasoning, we select correctly classified sentences with highlight the keywords that have high attention weights. Then, we calculate the top frequent words for each category to demonstrate whether the attention mechanism is valid to identify the important words. This provides the details on the capability of the proposed system on analyzing the clinical notes, and meanwhile allowing us to automatically extract keywords of sentences that are most relevant to the corresponding category.

The rest of the paper is organized as follows: Section II presents related work; system design and models are detailed in Section III; Section IV provides the data set description; Experimental results are given in Section V; Section VI concludes the work and list some future work. 

\section{Background and Related Work}

A wide variety of machine learning models have been applied to text document classification such as k-Nearest-Neighbor \cite{rogati2002high}, support vector machines (SVMs) \cite{joachims1999transductive}, convolutional neural networks (CNNs) \cite{conneau2016very} and recurrent neural networks (RNNs) \cite{liu2016recurrent}. In the clinical domain, document classification algorithms have been used to predict cancer stage information in clinical records \cite{yim2017classification}, to classify radiology reports by using ICD-9 code \cite{garla2012knowledge} and to classify whether a patient has psychological stress or no \cite{winata2018attention}. The most recent research shows that categorizing `free text' clinical notes is often related to other tasks of analyzing the content of Electronic Health Record (EHR) for decision support. The tasks include information extraction and information representation generation. Information extraction often refers to biomedical concept and event extraction, such as extracting gene expression \cite{zhu2017gram}, symptoms \cite{iyer2018incorporating}, diseases (including abbreviations) \cite{li2017neural} and drug to drug interaction \cite{zhao2016drug} \cite{xu2018leveraging} and so on. Other text analyses and NLP applications in the clinical field are relevant to clinical outcome prediction  \cite{rumshisky2016predicting}.

In recent years, the distributed representation of words or concepts which is called embedding gained interest in the research areas of text mining, natural language processing, and health informatics \cite{Mikolov:2013} \cite{moen2013distributional} \cite{tulkens2016using}. The embedding has been studied for biomedical text classification, clustering \cite{tulkens2016using} \cite{Zhu:2017} and biomedical entity extraction \cite{yadav2017entity}, where a word is a basic unit for the text documents and the word embedding is learned through neural networks including CNN or LSTM. The most recent text embedding is BERT \cite{devlin2018bert} which consists of a multi-layer bidirectional Transformer encoder. BERT process text as tokens since it is trained on unsupervised tasks to predict masked tokens. To the best of the authors' knowledge, BERT has not been investigated for clinical text classification and keyword extraction.

Both LSTM and Bidirectional LSTM has been widely used to for biomedical or chemical entity recognition  and extraction from the textual data \cite{luo2017attention} \cite{wang2018cross} \cite{xu2018improving}. However, it is rarely seen that LSTM or BiLSTM has been used for clinical text classification and keywords extraction. Although the most recent research used BERT and BiLSTM for medical text inference \cite{lee2019ncuee}.

One of the most recent research used the attention mechanism with LSTM to predict daily sepsis, myocardial infarction (MI), and vancomycin antibiotic administration through analyzing patients' ICU data in the MIMIC-III data set \cite{kaji2019attention}. This research demonstrates that the attention mechanism can extract the influential input variables that are related to the predictions.  Different from the previous study where the attention mechanism is used for classifying data, we aim to extract the keywords in the text that drive the classification using the attention mechanism. The attention mechanism approach enables human interpretable clinical text classification.

\section{System Design and Models}

\begin{figure*}[htbp]
\centerline{  \includegraphics [scale =0.6] {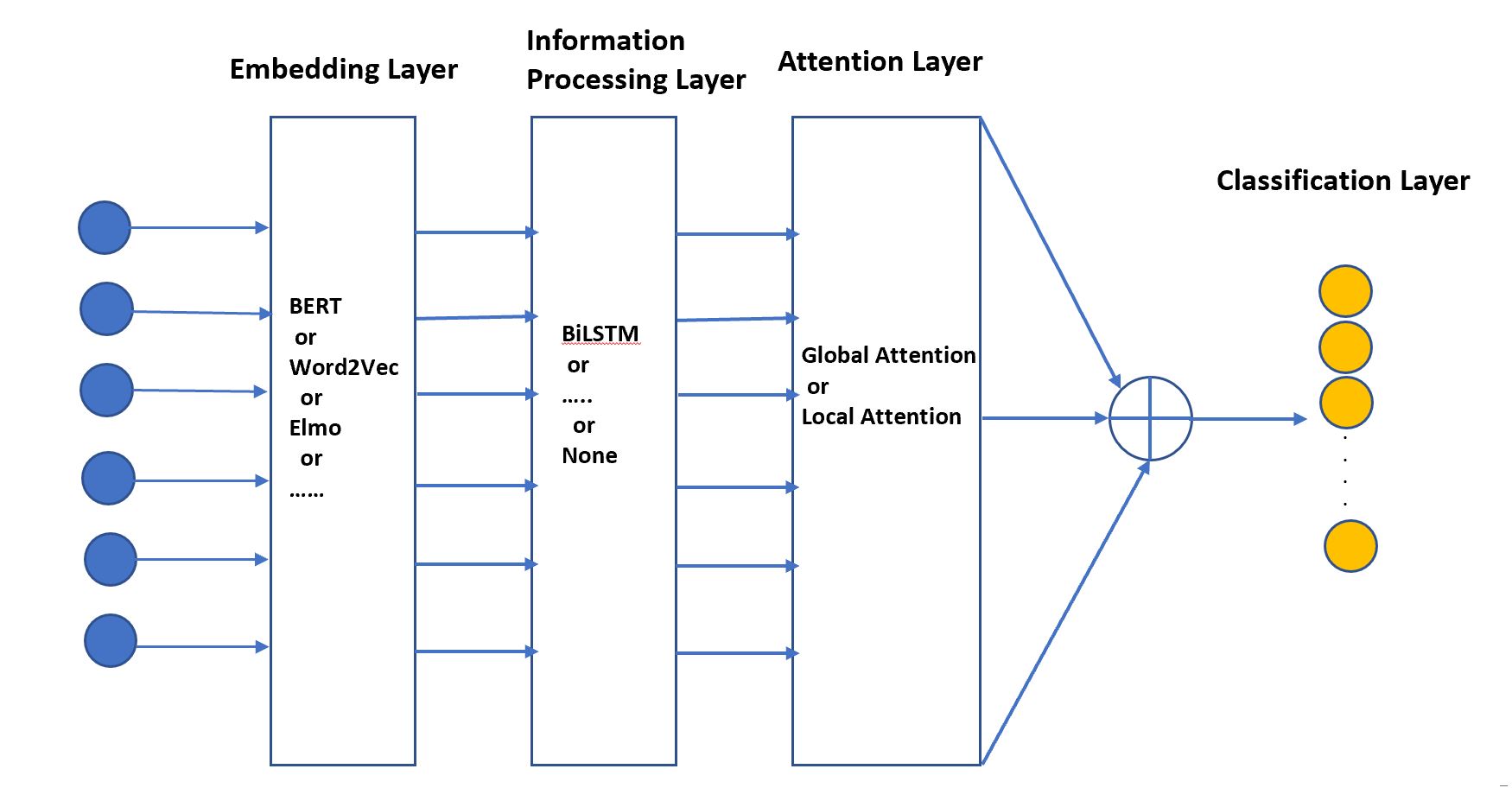}}
\caption{Attention Based Model System Design}
\label{fig:modelarchitect}
\end{figure*}

In this research, we develop and evaluate the neural attention mechanism with two different embeddings and text classification strategies. The attention layer is employed for each model before the classification layer to extract the keywords that highly influence the classification. Figure \ref{fig:modelarchitect} shows our system design. The input to the system is words or tokens. The first layer is to convert the word to embeddings. The second layer is the information processing layer, which can be a designed neural network, such as BiLSTM to process in the embeddings. The embedding layers can be BERT, Word2Vec or other embeddings which fits more the application domain. The attention layer is before the classification layer, which is used to identify the importance of the words for text classification. The following subsection presents the details of the text embedding, text classification models, and attention mechanism for keywords extraction.

\subsection{BERT}

The BERT model architecture is based on a multi-layer Transformer encoder, which was originally implemented by Vaswani et al.  \cite{vaswani2017attention}. Devlin et al. \cite{devlin2018bert} introduced the BERT Transformer based on using bidirectional self-attention. This bidirectional mechanism removes the restrictions that self-attention can only incorporate the context from one side: left or right. Different from other embedding generation architecture, such as Word2Vec \cite{Mikolov:2013}, the input to the BERT model are not vectors that represent words. Instead, the input includes token, segment, and position embeddings. The token embedding is WordPiece embeddings \cite{wu2016google} that contains 30k tokens. 

The base BERT model is pre-trained using two unsupervised tasks: (1) Masked Language Model (LM) - a task to predict some random masked tokens in the input. The objective is to train bidirectional encoder. (2) Next Sentence Prediction (NSP) - a task to predict the following sentence of the input sentence. The objective is to understand sentence relationships, so that the pre-trained BERT model can be better fit to other NLP applications, such as Question Answering (QA) and Natural Language Inference (NLI) where sentence relationships are crucial. In this research, we make use of the base BERT model that is in TensorFlow Hub \cite{url_bert}. It has 12 transformer blocks, 12 self-attention heads, and the hidden size of 768. 

The BERT base model can be fine-tuned for text classification by simply adding a softmax classification layer on top of the BERT model to predict the class $c$ of a given text sequence, as Equation \ref{softmax1}. The input to the softmax layer is the last hidden layer output H of the first token that represents the original text sequence.

\begin{equation}\label{softmax1}
p(c|H) = softmax(WH)
\end{equation}

where $W$ are parameters for the classification layer. They are fine-tuned with all the parameters from BERT to maximize the log-probability of the correct label. 

\subsection{BERT with Attention Layer}
Although the fine-tuned base BERT model can be used for text classification, however, it is not easy to extract the keywords that drive the decision process of the model. Hence, we propose adding an attention layer before to capture the attention of the neural network on each token. With base BERT model, there are two output options to connect the model with a specific language task. One is sentence level output, the other is token level output. In this research, the attention layer is built on top of the token level output. The embedding representation of the tokens then concatenated into vectors to present a document. The number of neurons of the attention layer is defined by the maximum number of tokens in the text collection. The output of the attention layer then connects to the classification layer by applying the $relu$ activation function. By applying the $relu$ activation function, the relationship between tokens and the output can be well captured. 

In this case, the attention weight of each token is defined by the $softmax$ of the output of the attention layer, which is used to identify the importance of tokens. It is worth noting that the attention weights are not directly applied to the classification layer.

\subsection{BiLSTM}
Long Short term memory(LSTM) network \cite{hochreiter1997long} is a type of Recurrent neural networks (RNN) which is capable of connecting the previous data input to perform the current operation. Unlike traditional RNN, LSTM contains four gates interacting with each other in different ways. Each LSTM cell considers three inputs: the current input, the preceding state and the output of the preceding state. LSTM network relies on the state of its cells, and the state of the cells are updated based on four gates: the forget gate which removes the irrelevant data that has been received from the preceding hidden state; the input gate which determines which values are to be updated; the input modulation gate is where a vector is created with new values known as candidate values are generated to be added to the current cell state later; and the output gate which determines what information to output. LSTM solves the vanishing gradient problem the traditional RNN. To extend its capabilities on connecting input from two direction, bi-directional LSTM (BiLSTM) was introduced \cite{huang2015bidirectional}. BiLSTM comprises of two independent LSTM networks to generate an output $h$ for a given input $x$. One network traverses the information from the past to future, known as forward pass $(\overrightarrow{h_x})$ and another network traverses the information from the future to the past, known as reverse pass $(\overleftarrow{h_x})$. The element-wise sum operation is used to combine the outputs of forward pass and backward pass, given as Equation \ref{bilstme}.

\begin{equation}\label{bilstme}
    h_x = \overrightarrow{h_x} \oplus \overleftarrow{h_x}
\end{equation}

BiLSTM has been used for text classification \cite{zhou2016text}. Typically, the input the BiLSTM are word embedding. The number of neurons in each layer is the maximum length of the input document measured by words. The bi-directional nature of the BiLSTM incorporates the context from both sides of an input word sequence.

\subsection{BiLSTM with Attention Layer}
In this research, we investigate the attention layer on top of the BiLSTM layer to capture important words that drive the decisions of the document classification. In the attention layer, we introduce an attention weight matrix $Z$ (Equation \ref{attb}) to calculate the relationships between the current target word and all words in the document. The attention weight is calculated as the weighted sum of output vectors of BiLSTM, $H = \{h_1,h_2,...h_T\}$. The $w^\intercal$ is the transpose of the trained parameter vector. The attention weight reflects the importance of the words for the classification output.
\begin{equation}\label{attb}
    Z = softmax (w^\intercal (\tanh{(H)}))
\end{equation}
The output of the attention layer and the output of the BiLSTM will then be used to calculate the context vector $CV$, which is defined as Equation \ref{outatt}, to feed to the classification layer which is used to predict the category of the input document.
\begin{equation}\label{outatt}
    CV = HZ^\intercal
\end{equation}

\subsection{Attention Weights and Keywords Extraction}
Through attention weights obtained from the attention layer, we extract the keywords from each input text documents. Given a sequence of attention weights $Att = \{att_1, ..., att_n\}$ obtained from an input document, first, we identify the word or token that has highest attention weight $att_{max}$. Then, we calculate the difference between the rest of the attentions to $att_{max}$. The value of the $n^{th}$ percentile of the difference can be used as a threshold (Equation \ref{atts}) to find the important words or tokens for the input document. The tokens are then combined into words. We notice that with BERT model, for some situation, not all tokens of a word have the same level of attention weights. In that situation, if one of the tokens has attention weight makes it pass the threshold, the whole word is extracted as keywords. For all implemented models in this research, $n$ is set to be 10. 

\begin{equation}\label{atts}
    att_{threshold} = Percentile((att_{max}-att_i), n)
\end{equation}

\section{Data Set Description}
\begin{figure}[htbp]\label{distributiondata}
\centerline{  \includegraphics [scale = 0.5] {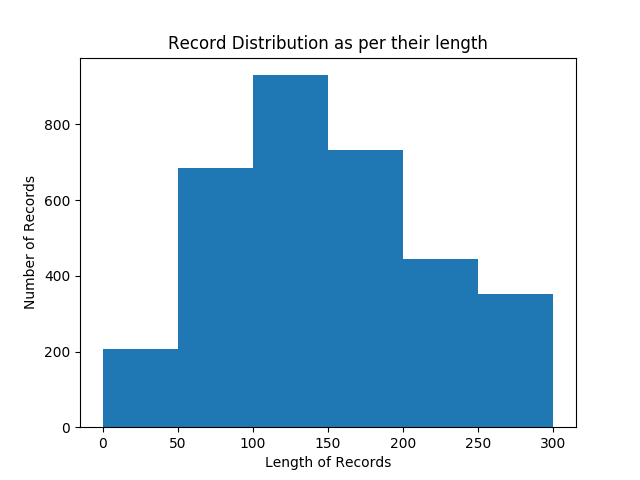}}
\caption{Length distribution of Progress Notes of our dataset}
\label{fig}
\end{figure}

The data set used in this study was extracted from a large academic medical center's EHR system.  In total, there are 3981 clinical progress notes extracted. These progress notes belong to 12 different categories. Table \ref{datad} shows the number of notes in each category. The progress notes are `free text' documents with many different sections written in them, such as `date', `patient name',`gender', `age', `medication', `allergies', `history of present illness' and so on. In this research, we use `history of present illness' section, which contains the most information as `free text' to demonstrate our system. The length of the `history of the present illness' section based on the number of words varies. However, the majority of the document contains less than 300 words for that section. Figure \ref{distributiondata} shows the distribution of the documents with less than 300 words by length. Since the base BERT model can only process documents with less than 512 tokens, we choose the documents with less than 250 words for this research.

\begin{table}[htb] \label{datad}
\setlength{\tabcolsep}{1.5pt}
\centering
\caption {Clinical Progress Notes in the Categories} \label{tabdistribution} 
\begin{tabular}{|c|c|c|c|}
\hline
Category & Document No. \\ \hline
Breast Care & 1965  \\ \hline
Urology & 98 \\ \hline
Bariatics & 33  \\ \hline
Dermatology & 75 \\ \hline
Endo-Diab & 263  \\ \hline
Geriatrics & 45  \\ \hline
GI-Gen & 55  \\ \hline
Nephrology & 48  \\ \hline
Orthopedics & 253  \\ \hline
Pain Management & 42  \\ \hline
Pulmonary & 86  \\ \hline
Sleep Med & 36  \\ \hline
\end{tabular}
\end{table}

\section{Experimental Results}
In this research, we implement three different models with the attention mechanism. One is BERT with attention layer (FT-BERT+Att), the other two is BiLSTM with attention layer using different embeddings. One uses a simple one-hot encoder embedding (OE+Att+BiLSTM), the other one uses pre-trained BERT based token embedding (PT-BERT+Att+BiLSTM). The reason to consider these two embeddings is to investigate how much impact of the embedding has on the performance of text classification and keyword extraction. One-hot encoder embedding captures very minimum semantic relationships between words, and the token embedding represents smaller units than words.

\subsection{Classification Accuracy}
Table \ref{table1} shows the classification accuracy of the implemented models.  We also compare the basic fine-tuned BERT without attention layer to demonstrate the state-of-the-art text classification using BERT. The results show that fine-tuned BERT model works better than the other BiLSTM based models. The fine-tuned BERT model with attention layer works the best. We also noticed that the high accuracy is gained mainly because of the accuracy on category Breast Care is high by using all three models. All models do not work well with the categories that have a smaller amount of instances, such as Bariatrics.

\begin{table}\label{table1}
\caption{Classification Accuracy of the Deep Learning Models}
\centering
\setlength{\tabcolsep}{3pt}
\begin{tabular}{|l|l|l|}
\hline
Models & Training & Test\\ \hline 
Basic Fine-Tuned BERT Text Classification & 99.8\% & 95.5\% \\ \hline
Fine-Tuned BERT+Attention Layer (FT-BERT+Att) & \textbf{99.9\%} & \textbf{97.6\%} \\ \hline
Pre-trained BERT+Attention Layer+BiLSTM & 95.6\% & 93.8\% \\ 
(PT-BERT+Att+BiLSTM) && \\\hline
One-hot Encoder+Attention Layer+BiLSTM & 90.2\% & 94.2\% \\ 
(OE+Att+BiLSTM) &  &  \\ \hline
\end{tabular}
\end{table}

\subsection{Keywords Extraction based on Attention Weights}
\subsubsection{Visualization of the Keywords with High Attention Weights}

\begin{figure}[htbp]
\centering
\includegraphics[scale=0.7]{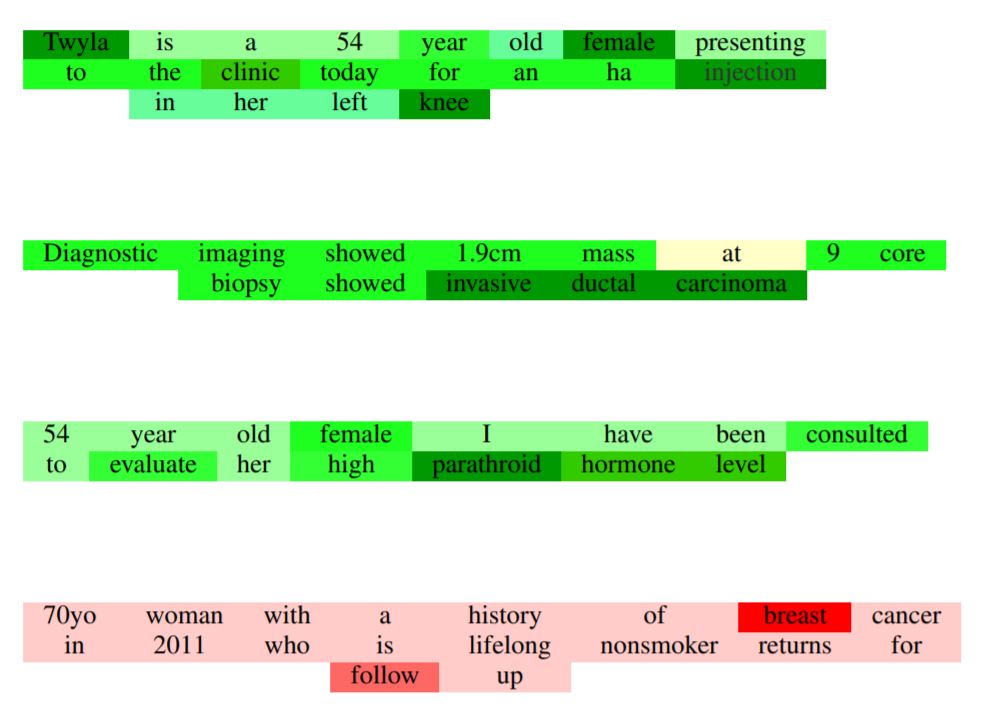}
\caption{Keywords with High Attentions Weights using FT-BERT+Att Model}
\label{fig_att_sent}
\end{figure}

\begin{figure}[htbp]
\centering
\includegraphics[scale=0.5]{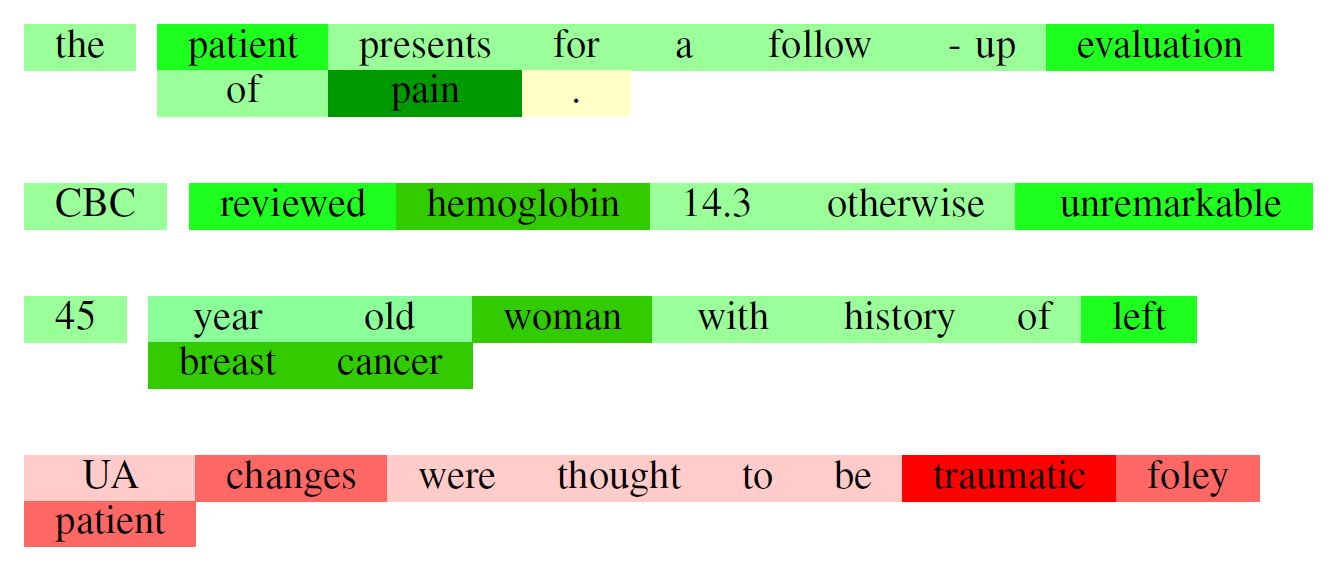}
\caption{Keywords with High Attentions Weights using PT-BERT+Att+BiLSTM Model}
\label{fig_att_sent1}
\end{figure}

\begin{figure}[htbp]
\centering
\includegraphics[scale=0.7]{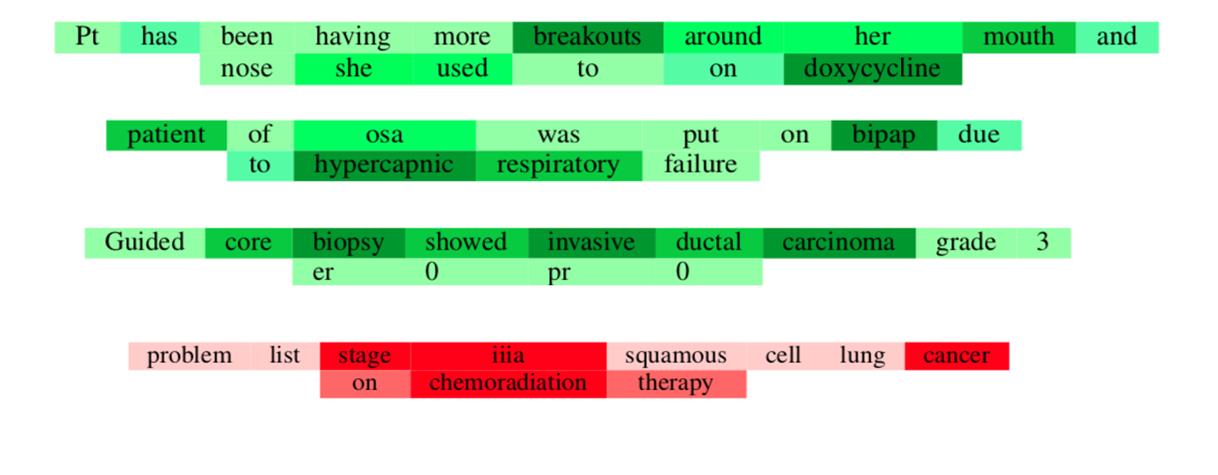}
\caption{Keywords with High Attentions Weights using OE+Att+BiLSTM Model}
\label{fig_att_sent2}
\end{figure}

Figure \ref{fig_att_sent} to Figure \ref{fig_att_sent2} shows the visualization of the keywords with attention weights over the specified the threshold in sentences with respect to the classification of the document by using FT-BERT+Att Model, PT-BERT+Att+BiLSTM model, and OE+Att+BiLSTM, respectively. Sentences in the documents that are correctly classified are colored green, while sentences that are misclassified is colored red. The colors corresponds to the attention weights calculated by the models. The darker the color is, the higher the weights are.  

Figure \ref{fig_att_sent} presents some examples of the sentence with the high attention words identified by the FT-BERT+Att Model. The first sentence is correctly classified to category Orthopedics, which is the branch of medicine dealing with the correction of deformities of bones or muscles. The most important keywords identified by attention weights are: `Twyla', `female', `injection', and `knee'. Word `knee' is very related to this category comparing to the other three words. `Twyla' is the name of the patient; however, its attention weight is relatively high in the document. We checked that among 253 documents, 13 of them mentioned this patient name in the `history of present illness' section. The frequency of it might cause its high attention weight. The second sentence is correctly classified to category Breast Care, which is the clinic offers breast care health, including screening, diagnosis, treatment options, symptom management, and so on. The words that have high attention weights are `invasive', `ductal', `carcinoma'. These words together is a specific type of breast cancer. The third sentence is correctly classified to category Endo-Diab, which is a branch of medicine that deals with hormones and glands that produce them. Only one word `parathyroid' has high attention weight, and it is a type of hormone which is usually seen by the physician of Endo-Diab. The last sentence, which is colored in red, is misclassified to Breast Care. The correct category should be Pulmonary. Based on the content and the identified keywords, we can tell that the attention word `breast' mislead the classification result.

In Figure \ref{fig_att_sent1} presents some examples of sentence with the high attention words identified by the PT-BERT+Att+BiLSTM model. The first sentence is correctly classified to category Pain Management, which is the branch of medicine that applies science to the reduction of pain. The most important keyword identified in this example is `pain', which is very related to this category. The second sentence is correctly classified to category Nephrology, which is a branch of medical science that deals with diseases of the kidneys. The word that has high attention weight is `hemoglobin' which is a type of blood test. It relates to chronic kidney disease (CKD). The third sentence is correctly classified to category Breast Care. The identified high attention words `woman' and `breast cancer' are very related to this category. The last example was misclassified to GI-Gen, the attention keywords are `changes', `traumatic', `foley', and `patient' which are not strongly related to the true category Dermatology.

Figure \ref{fig_att_sent2} presents some examples of sentence with the high attention words identified by OE+Att+BiLSTM Model.   The first sentence is correctly classified to category Dermatology, which is the branch of medicine conducts clinical and basic investigations of skin biology and researches the diagnosis and treatment of skin disease. The most important keywords identified by attention weights are: `breakouts', `mouth', and `doxycycline'. Word `doxycycline' is the type of medicine being used to treat many different bacterial infections, such as acne. The `breakouts' and `month' correlate to `doxycycline' in this case, so they also have high attention weights. The second sentence is correctly classified to category Sleep Medicine is the medical branch devotes to the diagnosis and therapy of sleep disturbances and disorders. The words that have high attention weights are `patient', `bipap', `hypercapinc', and `respiratory'. The `bipap' is a sleep apnea treatment. The `hypercapnic respiratory failure' is related to `bipap' in this case. The third sentence is correctly classified to category Breast Care. The high attention words, `biopsy', `invasive', and `carcinoma' are all highly related to breast cancer. The last sentence, which is colored in red, is misclassified to Endo Diab. The correct category should be Pulmonary. Although the content has `lung cancer' and somehow related to Pulmonary, however, `lung' is not captured as attention word in this case. Hence, it is misclassified.

\begin{table*}[]\label{top101}
\centering
\caption{Top 10 Frequent Words of Categories: Breast Care, Urology, Bariatics, Dermatology, Endo-Diab, Geriatrics}
\begin{tabular}{|c|c|c|c|c|c|c|}
\hline
Methods & \multicolumn{6}{c|}{Categories} \\\cline{2-7} 
& Breast Care & Urology & Bariatics & Dermatology & Endo-Diab & Geriatrics \\\hline
FT-BERT+Att & carcinoma &patient  &levels&old&preg&today\\
& breast & urinary &gas&history&last&pain\\
& negative &today &diet&allergies&patient&staff\\
& left & ambulatory &states&presents&diabetes&day\\
& showed & hematuria  &weight&breast&follow&presents\\
& patient & seen  &well&none&cancer&daily\\
& biopsy &procedure  &food&oral&been&sleep\\
& carcinomy & states &walking&skin&hyper&headache\\
& invasive & year  &pain&past&thyroid&nursing\\
& mass &report  &vomiting&right&met&old\\
\hline
PT-BERT& breast &  states &food&skin&type&pain\\
+Att+BiLSTM & left & prior& diet &history&breast&thigh\\
& ductal&  urinary &vitamins &states&cancer&neuropathy\\
& mastectomy & last &  problems&cancer&visit&leg\\
& chemotherapy& cystoscopy&bowel &lesion&medication&hurt\\
& completed & atb&protein &breast&high&concern\\
& underwent &reports &regarding &derm&metformin&area\\
& biopsy & nocturia  &levels&upper&surgery&tigan\\
& node &urine &walking&thigh&mcg&sinusutitis\\
\hline
OE+Att+BiLSTM & breast & old &numbness&old &old&wants\\
&  pr & retention &tingling&diabetes&diabetes&everplus\\
&  carcinoma  & year &bilateral&years&year&ability\\
& mastectomy & male  &mostly&presents&hypothyroidism&300mg\\
&  negative & taken &recurrent&abdomen&follow&walk\\
& biopsy & urgency &hand&months&type&feet\\
&  grade  & urinary &symptoms&keratosis&symptoms&rehab\\
& positive  & performing  &years&underwent&patient&feels\\
& mass &  cancer&years&clinic&back&know\\
& retention &resident  &find&female&patient&progress\\
\hline
\end{tabular}
\end{table*}

\begin{table*}[]\label{top102}

\caption{Top 10 Frequent Words of Categories: GI-Gen, Nephrology, Orthopedics, Pain Mgmt, Pulmonary, Sleep Med }
\centering
\begin{tabular}{|c|c|c|c|c|c|c|}
\hline
Methods & \multicolumn{6}{c|}{Categories} \\\cline{2-7} 
& GI-Gen & Nephrology & Orthopedics & Pain Mgmt& Pulmonary & Sleep Med \\\hline
FT-BERT+Att & colonoscopy  & urine &pain&pain&copd&sleep\\
&diarrhea  & uropathy &patient&follow&history&chemotx\\
& pain & originally&left&rated&last&apnea\\
& dysphagia & history &state&initial&follow&epworth\\
& recreational &pleasant   &knee&bilateral&returns&polyp\\
& colon & year  &well&today&old&sleepiness\\
& vomiting & colon &denies&treatment&visit&better\\
& ago & cancer &right&factors&today&patient\\
& negative &  symptoms &fracture&spasms&denies&time\\
& past & mild &returns&walking&cough&night\\
\hline
PT-BERT & pain & creatinine  &pain&severity&female&cpap\\
+Att+BiLSTM & colonoscopy & reviewed  &female&pain&sleep&quality\\
& denied & nonsteroidal &knee&fluctuates&cough&use\\
&cancer & hemoglobin &numbness&weather&cancer&osa\\
& vomiting & urinary  &fracture&treatment&use&download\\
& showed& sodium   &presenting&mod&winded&uses\\
& nausea &  cancer&last&bilaterally&shortness&used\\
&diarrhea  & urosepsis &symptoms&factors&obstructive&sputum\\
&fever  & ureter  &presents&extremity&dyspnea&pressure\\

\hline
OE+Att+BiLSTM & abdominal  &old&old&pain&old&takes\\
&never   &september&denies&severe&x2&contiue\\
&discomfort   & ultrasound &female&moderate&cancer&sleepy\\
&diabetes    &a1c  &today&mod&coil&due\\
&colonoscopy & kidney  &reports&&using&experience\\
&chills   & previous &presenting&&cabg&narcotics\\
&fine&initial    &follow&&chemotherapy&lots\\
&    & visit &right&&diabetes&back\\
&   & routine  &year&&time&\\
&  &bilateral  &quite&&chest&\\
&  & potassium &improved&&returns&\\
\hline
\end{tabular}
\end{table*}
\subsubsection{Frequent Keywords of Each Category}
Based on the example analysis, we can tell that the attention weight-based keywords extraction might also extract some words that are not directly related to the category. Each model can include different words that are not directly related to the category.  So, we investigate the top frequent keywords of each category by using these three models. Table II and III show the top 10 frequent words of the different categories after removing the stop words. For the category Breast Care, which has the most number of documents, there 6 to 7 of the frequent words identified by each model are directly related to the category.  There are some keywords identified by all three models, such as `breast', `carcinoma', and `biopsy'. 

The results show that for other categories, only 1 to 3 words are identified by all three models, and those words are normally highly related to the corresponding category, such as `urinary' for category Urology, `skin' for Dermatology and so on. It is found that there are more overlapped keywords between FT-BERT+Att model and PT-BERT+Att+BiLSTM model. The reason could be that they both based on the token embeddings. The OE+Att+BiLSTM model captures much less related keywords than the other two models for most of the categories. For some category, such as Bariatrics, OE+Att+BiLSTM model does not capture any directly related words to the category. The classification accuracy for that category is very low as 33\%. On the other hand, after applying the attention threshold and removing the stop words, OE+Att+BiLSTM model can't identify ten frequent words for some categories, such as Pain Management category. We hypothesize this is because the one-hot encoder embedding captures no semantic relationships between different words. Hence, the decision is often based on the repetition of the same words in the category.  For example, `old' occurs many times in different categories; it is identified as a keyword with high attention weight for different categories.

\section{Discussion}
Both classification and keyword extraction results demonstrate that the attention-based deep learning models are capable of clinical text classification. Without using the attention layer, fine-tuned BERT model can also achieve high accuracy in classification. With attention layer, the fine-tuned BERT performs better than the other models on text classification. The objective of the attention layer is to extract the keywords or phrases to interpret the decision process of the network for text classification. The selected sentences in Figures \ref{fig_att_sent} to \ref{fig_att_sent2} demonstrate the visualization of the important words through calculating the attention weights. Often, we find that when the important words are identified correctly, the classification results are also correct. This shows that the attention layer for the text classification interpretation is effective. 

The attention-based models also demonstrate that different embedding mechanisms and classification mechanisms can lead to different results. The capture keywords are different. Some of them work better than the other. Based on the identified frequent keywords of the categories, we conclude that the embedding layer is crucial for the text classification and keyword extraction. We explore the token-based embedding and simple one-hot embedding in this paper. We expect to explore other embeddings in the future, especially the pre-trained embeddings using the biomedical data set, such as Clinical BERT \cite{alsentzer2019publicly} and BioWord2Vec \cite{zhang2019biowordvec}.

\section{Conclusion and Future Work}
In this paper, we examine three attention-based deep learning models for clinical progress notes classification and keywords extraction. Two of the models based on the most recent language embedding model - BERT, the other one based on a simple one-hot encoder embedding. Although three models gain good performance on progress notes classification, through the attention layer in the three models, we are capable of interpreting the text classification process of the models. Words with high attention weights are the important words that associate with the text categories. This research presents interpretable models for text classification and demonstrates the power of the attention-based approach for model interpretability and evaluation.

The future work includes evaluating the model on different embeddings and considering building attention-based model by incorporating syntactic relationships between words for keyphrases extraction and interpretation. 

\section*{Acknowledgment}
This research was support by IU Health, Department of CIT
at IUPUI, with funding from National Science Foundation and
United States Department of Defense. The authors would also
like to thank Dr. Feng Li and Sheila Walter for their support.

\bibliographystyle{plain}
\bibliography{bibm}

\end{document}